\documentclass[conference]{IEEEtran}

\usepackage{amsmath,amssymb,amsfonts}
\usepackage{bm}
\usepackage{graphicx}
\usepackage{xcolor}
\usepackage{hyperref}
\usepackage{cite}
\usepackage{algorithm}
\usepackage{algorithmic}
\usepackage{tikz}
\usepackage[section]{placeins}
\usetikzlibrary{arrows.meta}


\newcommand{\owl}{\textit{OWL}}

\newcommand{\er}{\hat{e}_r}
\newcommand{\bw}{\boldsymbol{\omega}}
\newcommand{\bt}{\mathbf{t}}
\newcommand{\br}{\mathbf{r}}
\newcommand{\bOmega}{\boldsymbol{\Omega}}

\begin{document}

\title{Segmentation of Moving Objects During Camera Motion Using the \owl{} Function}

\author{
  \IEEEauthorblockN{Daniel Raviv \quad Juan D.~Yepes}
  \IEEEauthorblockA{Electrical Engineering and Computer Science Department\\
    Florida Atlantic University, Boca Raton, Florida, USA\\
    \{ravivd, jyepes\}@fau.edu}
}

\maketitle

\begin{abstract}
We present a principled, analytical framework for segmenting multiple
independently moving rigid objects during single camera motion, built directly
upon the \owl{} perception function~\cite{raviv2026owl}.
\owl{} encodes two fundamental visual motion cues, perceived looming~$L$
and perceived rotation~$\omega$, into a single closed-form representation
of relative 3D structure.
A key consequence of this structure is that the unit direction of relative
translation~$\boldsymbol{\hat{t}}$ is analytically recoverable per image point from~$L$
and~$\omega$ alone.
All points belonging to a non-rotating rigid body share the same
$\boldsymbol{\hat{t}}$, whereas points on different rigid bodies generally yield distinct values.
This furnishes a natural segmentation criterion: clustering per-point
$\boldsymbol{\hat{t}}$ estimates partitions the visible scene into independently moving
rigid bodies.
Building on the 2D complex form of \owl{}, we derive the framework in 3D
using quaternions, for a single moving camera.
The approach is parallel, camera rotation-invariant, and requires no
training data, no depth information, and no semantic priors.
\end{abstract}

\begin{IEEEkeywords}
motion segmentation, rigid body motion, visual looming, \owl{} function,
Segment Anything, quaternions, autonomous navigation, robotics.
\end{IEEEkeywords}

\section{Introduction}
\label{sec:intro}

\subsection{Conceptual Explanation}

A central challenge in autonomous navigation and robotics is the
real-time identification and separation of multiple independently moving
objects in a dynamic scene, without prior knowledge of the environment,
without depth information, and without trained models.

The \owl{} function~\cite{raviv2026owl} establishes that two instantaneous
per-point visual motion cues, perceived looming~$L$ and perceived
rotation~$\omega$, together encode the complete ratio between the relative
translational velocity and range from a single camera to any visible 3D
point.
This ratio can be computed from the image sequence, together with the
camera's rotation rate, without any prior knowledge of the scene.

The present paper develops a further consequence of this mathematical
structure: the unit direction of relative translation~$\boldsymbol{\hat{t}}$ is
analytically recoverable per point from~$L$ and~$\omega$ alone.
Points on the same rigid body share the same~$\boldsymbol{\hat{t}}$; points on
different rigid bodies produce generally different values.
This is the segmentation foundation developed here.

For each trackable image point, $L$ and the magnitude of~$\omega$ fix the
angle~$\alpha$ between the relative translation direction~$\boldsymbol{\hat{t}}$ and
the line of sight to that point, so~$\boldsymbol{\hat{t}}$ lies on a cone around
that line of sight.
Points on the same rigid body produce cones that share a common
direction,~$\boldsymbol{\hat{t}}$, which gives an intuitive geometric picture.
When the full perceived rotation vector is available, as it is from
de-rotated optical flow, a closed-form quaternion expression recovers
$\boldsymbol{\hat{t}}$ from each point independently, and segmentation becomes a
clustering problem on the unit sphere.

\owl{} stands in contrast to learning-based and foundation-model segmentation
approaches, including the Segment Anything Model
(SAM)~\cite{kirillov2023sam} and its video extension SAM~2~\cite{ravi2024sam2}, which
require large-scale training data and provide no analytical relationship
between visual motion and scene geometry.
Once objects are segmented by~$\boldsymbol{\hat{t}}$, the \owl{} representation or its 3D extension RoT can
be applied per segment to recover a scaled 3D point cloud for each
independently moving body.

\section{Literature Review}
\label{sec:lit}

\subsection{Motion Segmentation Methods}

Motion segmentation has a long history in computer vision.
Classical factorization methods~\cite{tomasi1992shape} decompose image
measurement matrices under affine camera assumptions.
Costeira and Kanade~\cite{costeira1998multibody} extended factorization to multiple
bodies.
The Generalized Principal Component Analysis (GPCA) of
Vidal~et~al.~\cite{vidal2005gpca} provides a polynomial algebraic framework for
segmenting subspace-structured data.
Sparse Subspace Clustering (SSC)~\cite{elhamifar2013ssc} exploits the
self-expressive property of unions of subspaces.
Spectral clustering~\cite{shi2000normalized} and graph-cut formulations group pixels
by motion similarity.
These methods fit global models, such as subspaces or affine
factorizations, jointly across many trajectories, and are sensitive to
noise and outliers.
\owl{}-based segmentation instead recovers~$\boldsymbol{\hat{t}}$ analytically and
independently at each point from the cues~$L$ and~$\omega$, with no global
model fitting.

\subsection{Optical Flow Methods}

The classical variational approach of Horn and Schunck~\cite{horn1981optical}
minimizes a global smoothness functional.
Lucas and Kanade~\cite{lucas1981iterative} introduced the local least-squares
approach.
Modern deep learning methods such as FlowNet~\cite{dosovitskiy2015flownet},
PWC-Net~\cite{sun2018pwcnet}, and RAFT~\cite{teed2020raft} achieve high
performance but require large labeled datasets and significant computational
resources.
A critical distinction emphasized in~\cite{raviv2026owl} is that~$\omega$, the
perceived rotation per image point, is not the same as the standard optical
flow vector.
The quantity~$\omega$ cancels the camera rotation contribution, making it
invariant to ego-rotation by construction.
\owl{}-based segmentation inherits this rotation invariance automatically.

\subsection{Semantic, Instance, and Panoptic Segmentation}

Semantic segmentation assigns a class label to every pixel.
Instance segmentation distinguishes individual object instances.
Mask R-CNN~\cite{he2017maskrcnn} established the dominant two-stage paradigm.
Mask2Former~\cite{cheng2022mask2former} unifies semantic and instance
segmentation using masked cross-attention.
Panoptic segmentation~\cite{kirillov2019panoptic} combines both tasks.
These methods are fundamentally category-driven: they distinguish objects
by what they are rather than by how they move, and cannot generalize to
categories unseen during training.
\owl{}-based segmentation is category-independent and motion-driven: any
independently moving rigid body is generally naturally segmented by~$\boldsymbol{\hat{t}}$, with no
concept of object class and no training required.

\subsection{Foundation Models for Segmentation}

SAM~\cite{kirillov2023sam} is a foundation model for promptable segmentation,
trained on over one billion masks, generating high-quality segmentation
masks in zero-shot fashion given a point, bounding box, or mask prompt.
SAM~2~\cite{ravi2024sam2} extends this paradigm to video via a streaming
transformer with a memory bank.
The contrast with \owl{}-based segmentation is fourfold.
First, SAM operates on appearance, not motion, so it cannot distinguish two
visually similar objects moving in different directions.
Second, SAM must be prompted, either by a user or by a dense grid of
points in its automatic mode, and its masks do not indicate which regions
move together, whereas \owl{}-based segmentation groups points by their
motion without prompts.
Third, SAM requires massive training infrastructure, whereas \owl{}-based
segmentation requires none.
Fourth, SAM produces 2D masks; \owl{}-based segmentation simultaneously
produces a scaled 3D point cloud per segment.
The methods are complementary: $\boldsymbol{\hat{t}}$ clusters can serve as
motion-grounded prompts for SAM, and SAM masks can provide spatial
regularization for \owl{} clustering.

\subsection{Biological and Ecological Perception}

The \owl{} framework draws direct inspiration from biological visual
systems.
Gibson~\cite{gibson2014ecological} introduced the ecological approach to visual
perception, emphasizing optic flow as the primary cue for self-motion and
object detection.
Neurons in the fly lobula plate are tuned to specific optic flow
patterns~\cite{krapp1996estimation}.
Looming-sensitive neurons in insects detect objects on collision courses
from expanding flow fields~\cite{srinivasan2004visual}, a direct biological
analog of the looming cue~$L$ in \owl{}.

The perceived rotation cue $\omega$ has direct biological correlates across
species. In the fly, FD (figure-detection) cells in the lobula plate are
specialized for relative motion between an object and the background, with
their selectivity shaped by inhibitory input from HS (horizontal system)
cells via CH interneurons, a circuit that effectively subtracts the global
rotational component from local motion signals, isolating the translational
residual~\cite{haag2002cite}.
Only the translational components of optic flow contain spatial
information, a principle the fly visual system enforces neurally by
separating rotation-encoding (HS/VS) cells from de-rotated translational
flow (FD cells), mirroring \owl{}'s distinction between camera rotation
$\Omega$ and perceived rotation $\omega$.
In primates, area MSTd encodes optic flow across a continuous spiral space
combining radial and rotational components~\cite{graziano1994cite}, with
the full spiral-space population providing more accurate self-motion
estimates than expansion-tuned neurons alone~\cite{layton2022cite},
consistent with the brain treating $\mathrm{ToR} = (L,\bw)$ as a
primitive joint representation rather than decomposing $L$ and $\omega$
into separate channels.

\section{Method}
\label{sec:method}

\subsection{Assumptions}
The following assumptions define the geometric scope of the framework.
\begin{enumerate}
  \item The scene consists of relatively moving objects, one of which can be the stationary environment.
  \item Each rigid body translates without rotating. The camera may both
        rotate and translate; its rotation is removed by de-rotation
        (Section~\ref{sec:derot}).
  \item Each rigid body (including the stationary environment) has a distinct unit translation direction~$\boldsymbol{\hat{t}}_k$
        relative to the camera.
  \item The relative translation is nonzero ($\bt \neq \mathbf{0}$), so that
        $L^2+\omega^2 > 0$.
\end{enumerate}

\subsection{\owl{} Fundamentals}

The \owl{} framework~\cite{raviv2026owl} establishes that two perceived visual
motion cues, looming~$L$ and perceived rotation~$\omega$, directly encode
the ratio of translational velocity to range between a moving camera and
any visible 3D point.
Given the relative translation vector~$\bt$ and the range
vector~$\br$ from the camera to a 3D point, the two scalar quantities are:
\begin{equation}
  L \;=\; \frac{\bt \cdot \er}{|\br|}
    \;=\; \frac{|\bt|\cos\alpha}{|\br|},
  \label{eq:L}
\end{equation}
\begin{equation}
  \omega \;=\; -\frac{|\bt|\sin\alpha}{|\br|},
  \label{eq:omega}
\end{equation}
where~$\alpha$ is the angle from~$\bt$ to~$\br$ measured
counterclockwise, and~$\er = \br/|\br|$ is the unit line-of-sight vector.
In 2D complex notation these combine as:
\begin{equation}
  \frac{\tilde{t}}{\tilde{r}} \;=\; L + j\omega.
  \label{eq:tor2d}
\end{equation}
This means that the ratio of translational velocity to range is directly
measurable from the two perceived cues, without requiring knowledge of
$\bt$ or $\br$ individually.
The magnitude satisfies $|\bt|/|\br| = \sqrt{L^2+\omega^2}$, enabling
scaled 3D reconstruction.

\subsection{Recovering \texorpdfstring{$\boldsymbol{\hat{t}}$}{t-hat}: The Segmentation Key}
\label{sec:tcap}

We now show how the unit translation direction~$\boldsymbol{\hat{t}}$ is recoverable
from~$L$, $\omega$, and the line-of-sight geometry.
This is the core analytical result that enables segmentation.

\subsubsection{Geometric Intuition}
\label{sec:intuition}

Consider a camera looking forward at points in front of it.
For each approaching point ($L>0$), the two cues fix the angle~$\alpha$
between the direction of motion~$\boldsymbol{\hat{t}}$ and the line of sight to
that point:
\begin{equation}
  \tan\alpha \;=\; \frac{|\omega|}{L}.
  \label{eq:alpha}
\end{equation}
Knowing only this angle, $\boldsymbol{\hat{t}}$ can be any direction that makes
angle~$\alpha$ with the line of sight; that is, $\boldsymbol{\hat{t}}$ lies on a
cone of half-angle~$\alpha$ around it.
A point lying exactly along the direction of motion has $\omega = 0$, and its
cone collapses to the line of sight itself; points farther from that
direction produce wider cones.
A second point on the same rigid body adds a second cone, and the two cones
intersect in at most two directions.
A third point, not collinear with the first two in the image, leaves a
single direction common to all three cones: $\boldsymbol{\hat{t}}$.

\subsubsection{3D Recovery from Quaternion Representation}

In 3D, the full angular velocity vector
\begin{equation}
  \bw \;=\; \frac{\er \times \bt}{|\br|}
  \label{eq:wvec}
\end{equation}
carries the orientation of the plane containing~$\bt$ and~$\br$.
Starting from the pure quaternion definitions~$T=(0,\bt)$
and~$R=(0,\br)$, the quaternion product $\mathrm{ToR} = T \otimes R^{-1}$
evaluates to:
\begin{equation}
  \mathrm{ToR} \;=\; T \otimes R^{-1} \;=\; (L,\;\bw),
  \label{eq:ToR}
\end{equation}
where the scalar part~$L = \bt\cdot\er/|\br|$ is the looming and the
vector part~$\bw = \er\times\bt/|\br|$ is the angular velocity vector.
Its magnitude $\omega = |\bw|$ satisfies
$\sqrt{L^2+\omega^2} = |\bt|/|\br|$.

From~$\bw$ and~$L$ the unit translation vector is recovered by a direct
quaternion multiplication.
Defining the unit quaternion $Q$:
\begin{equation}
  Q \;=\; \frac{(L,\,\bw)}{\sqrt{L^2+\omega^2}},
  \label{eq:Q}
\end{equation}
and the unit pure quaternion~$\hat{R} = (0,\er)$, the translation direction
follows from a single left-multiplication (not a sandwich product):
\begin{equation}
  (0,\,\boldsymbol{\hat{t}}) \;=\; Q \otimes \hat{R}.
  \label{eq:tcap_quat}
\end{equation}
Expanding~\eqref{eq:tcap_quat} explicitly using the quaternion product rule
$(p_0,\mathbf{p})\otimes(q_0,\mathbf{q})=(p_0 q_0 - \mathbf{p}\cdot\mathbf{q},\;p_0\mathbf{q}+q_0\mathbf{p}+\mathbf{p}\times\mathbf{q})$
yields (an equivalent vector-algebra derivation, without quaternions, is given in Appendix~\ref{app:vector_derivation}):
\begin{equation}
  \boxed{\boldsymbol{\hat{t}} \;=\; \frac{L\,\er \;-\; \er\times\bw}{\sqrt{L^2+\omega^2}}}.
  \label{eq:tcap_explicit}
\end{equation}

Equation~\eqref{eq:tcap_explicit} shows that~$\boldsymbol{\hat{t}}$ is fully determined
by the measurable quantities~$L$, $\bw$, and~$\er$, from a single point.
No scene depth or camera translation magnitude is required; only the
line-of-sight direction~$\er$, which follows from the pixel location and
the camera intrinsics.

The RoT quaternion, the 3D counterpart of \owl{} and the reciprocal of ToR,
takes the form:
\begin{equation}
  \mathrm{RoT} \;=\; \mathrm{ToR}^{-1}
    \;=\; \frac{(L,\,-\bw)}{L^2+\omega^2},
  \label{eq:RoT}
\end{equation}
and represents real 3D space scaled by~$1/|\bt|$; multiplying the entire
RoT visualization space by~$|\bt|$ (if known) recovers the actual 3D positions.

\subsubsection{How Many Points Are Needed?}

The number of points needed to uniquely determine~$\boldsymbol{\hat{t}}$ depends on what
is measured per point.

\textbf{Case~1: Scalar $L$ and vector $\bw$ available.}
When the full angular velocity vector~$\bw$ is known,
equation~\eqref{eq:tcap_explicit} resolves~$\boldsymbol{\hat{t}}$ uniquely from a single
point. The vector~$\bw$ carries two constraints per point: the
angle~$\alpha$ and the orientation of the $\alpha$-plane.

\textbf{Case~2: Scalars $L$ and $\omega$ available.}
Each point fixes only the cone of Section~\ref{sec:intuition}; three
non-collinear points are needed to single out~$\boldsymbol{\hat{t}}$.
This case serves as geometric intuition; the segmentation in
Section~\ref{sec:seg} uses Case~1, which operates per point. This is summarized in Table~\ref{tab:points}.

\begin{table}[ht]
\caption{Points Needed to Recover~$\boldsymbol{\hat{t}}$}
\label{tab:points}
\centering\small
\begin{tabular}{lcc}
\hline
Known per point & Constraints & Points needed \\
\hline
$L$ and $\bw$ (vector) & 2 (angle + plane) & 1 \\
$L$ and $\omega$ (scalars) & 1 (angle only)  & 3 \\
\hline
\end{tabular}
\end{table}

\subsection{Obtaining \texorpdfstring{$\bw$}{w} from De-rotated Optical Flow}
\label{sec:derot}

The angular velocity vector~$\bw$ can be obtained either by direct
measurement or by de-rotating the observed optical flow.
In the general 3D spherical formulation, the angular components of the
motion field satisfy~\cite{raviv2026owl}:
\begin{align}
  \dot\theta\cos\phi &= -(\omega_\phi + \Omega_\phi), \label{eq:flow3d_a}\\
  \dot\phi             &= \omega_\theta + \Omega_\theta, \label{eq:flow3d_b}
\end{align}
In vector form these combine as:
\begin{equation}
  \er \times \bw \;=\; \mathbf{f} \;+\; \bOmega \times \er,
  \label{eq:w_from_flow}
\end{equation}
where $\theta$ and $\phi$ are the azimuth and elevation of the line of
sight, $(\er,\hat{e}_\theta,\hat{e}_\phi)$ is the corresponding right-handed
spherical frame, $\omega_\theta,\omega_\phi$ and $\Omega_\theta,\Omega_\phi$
are the components of~$\bw$ and~$\bOmega$ in that frame, and
$\mathbf{f} = \dot\theta\cos\phi\,\hat{e}_\theta + \dot\phi\,\hat{e}_\phi$
is the angular optical flow vector.
Since~\eqref{eq:tcap_explicit} depends on~$\bw$ only through
$\er\times\bw$, substituting~\eqref{eq:w_from_flow} gives~$\boldsymbol{\hat{t}}$
directly from the measured flow and camera rotation:
\begin{equation}
  \boldsymbol{\hat{t}} \;=\;
  \frac{L\,\er \;-\; \mathbf{f} \;-\; \bOmega\times\er}{\sqrt{L^2+\omega^2}}.
  \label{eq:tcap_flow}
\end{equation}
In the 2D special case ($\phi = 0$), \eqref{eq:w_from_flow} reduces to the scalar identity:
\begin{equation}
  \omega \;+\; \Omega \;+\; \dot\theta \;=\; 0,
  \label{eq:derot2d}
\end{equation}
so that $\omega = -(\dot\theta + \Omega)$ is obtained directly from the
observed optical flow and camera rotation.
This shows that~$\bw$ (and thus~$\omega$) requires no specialized sensors
beyond a calibrated IMU or gyroscope providing~$\Omega$, together with
the image sequence providing~$\dot\theta$.

\subsection{Segmentation by \texorpdfstring{$\boldsymbol{\hat{t}}$}{t-hat} Clustering}
\label{sec:seg}

Given per-point~$\boldsymbol{\hat{t}}$ estimates recovered via~\eqref{eq:tcap_explicit},
segmentation proceeds by clustering: all points sharing the same~$\boldsymbol{\hat{t}}$ generally 
belong to the same rigid body.
In practice, since the theoretical assumption of purely translational
motion per object holds only approximately,~$\boldsymbol{\hat{t}}$ estimates from the same
body form a tight cluster on the unit sphere, while estimates from
different bodies form well-separated clusters.
The number of clusters generally corresponds to the number of independently moving
bodies in the scene, including the stationary background.

Any clustering method on the unit sphere can be used. In the results of
Section~IV, a pixel is labeled as moving (white in the binary masks) when
$\hat{\mathbf{t}} \cdot \hat{\mathbf{t}}_{bg} < 0.999$, i.e., when its
direction deviates from the background direction $\hat{\mathbf{t}}_{bg}$
by more than about $2.6^\circ$.

Since~$\boldsymbol{\hat{t}}$ is computed independently per point (requiring only
local~$L$ and~$\bw$ values), the entire pipeline is fully
parallel and real-time compatible.
Camera rotation is compensated by construction through the de-rotation
of the optical flow, so the segmentation result is independent of camera
ego-rotation.
Camera translation, by contrast, enters every~$\boldsymbol{\hat{t}}$, since
$\boldsymbol{\hat{t}}$ is the direction of the camera's translation relative to
each body.  It should be noted that rigid bodies whose relative velocities are parallel and point the
same way share the same~$\boldsymbol{\hat{t}}$ and cannot be separated by this
criterion alone; Assumption~3 excludes this case.
\section{Initial Simulation Results}
\label{sec:sim}

The following figures illustrate the key concepts of the framework across a
range of motion configurations, from a stationary scene under camera
translation to multiple independently moving, self-rotating objects under
simultaneous camera rotation and translation. In every scenario, $L$,
$\omega_\theta$, and $\omega_\phi$ are the analytical ground-truth cues
computed directly from the Unity simulation's known camera and scene motion,
rather than values estimated from image derivatives. The figures therefore
demonstrate the closed-form correctness of the $\hat{\mathbf{t}}$ recovery and
clustering geometry described in Section~III-C and Section~III-E.

In all figures, $\hat{\mathbf{t}}$ is rendered through a fixed continuous
mapping from direction to color, so pixels sharing the same $\hat{\mathbf{t}}$
share the same color.

\subsection{Stationary Scene}
\begin{figure*}[t]
\centering
\includegraphics[width=0.85\textwidth]{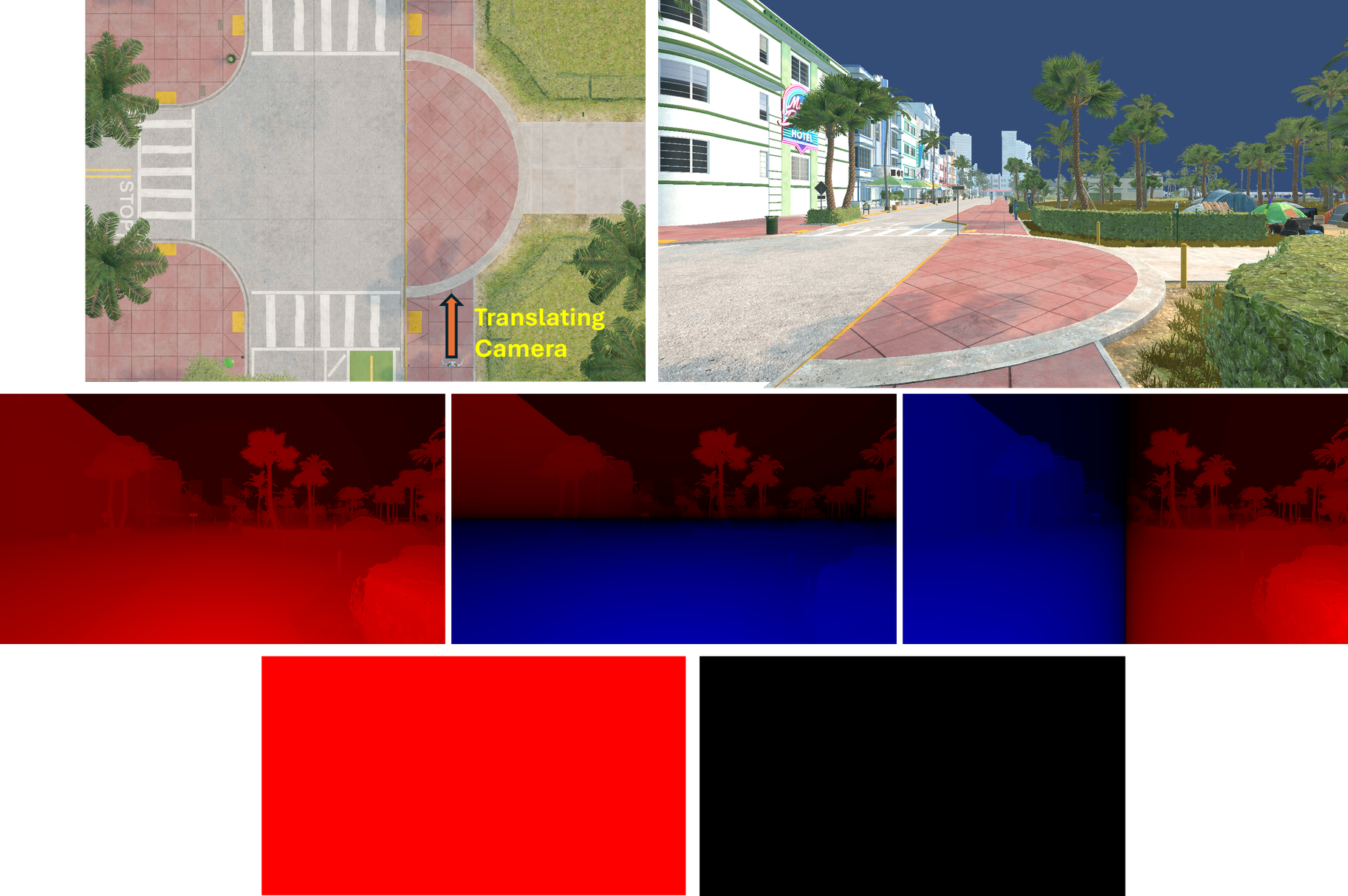}
\caption{Camera translating through a static environment (with no camera
rotation). Top row: a top-down schematic of the camera motion and a
representative rendered frame. Middle row, left to right: the ground-truth
cues $L$, $\omega_\theta$, and $\omega_\phi$. Bottom row: the recovered
$\hat{\mathbf{t}}$ field visualized as RGB color. Since
the scene is a single rigid body undergoing pure camera-relative
translation, $\hat{\mathbf{t}}$ collapses to a single color (red)
everywhere, as predicted in Section~\ref{sec:tcap}.}
\label{fig:stationary_qual}
\end{figure*}

\begin{figure*}[t]
\centering
\includegraphics[width=0.85\textwidth]{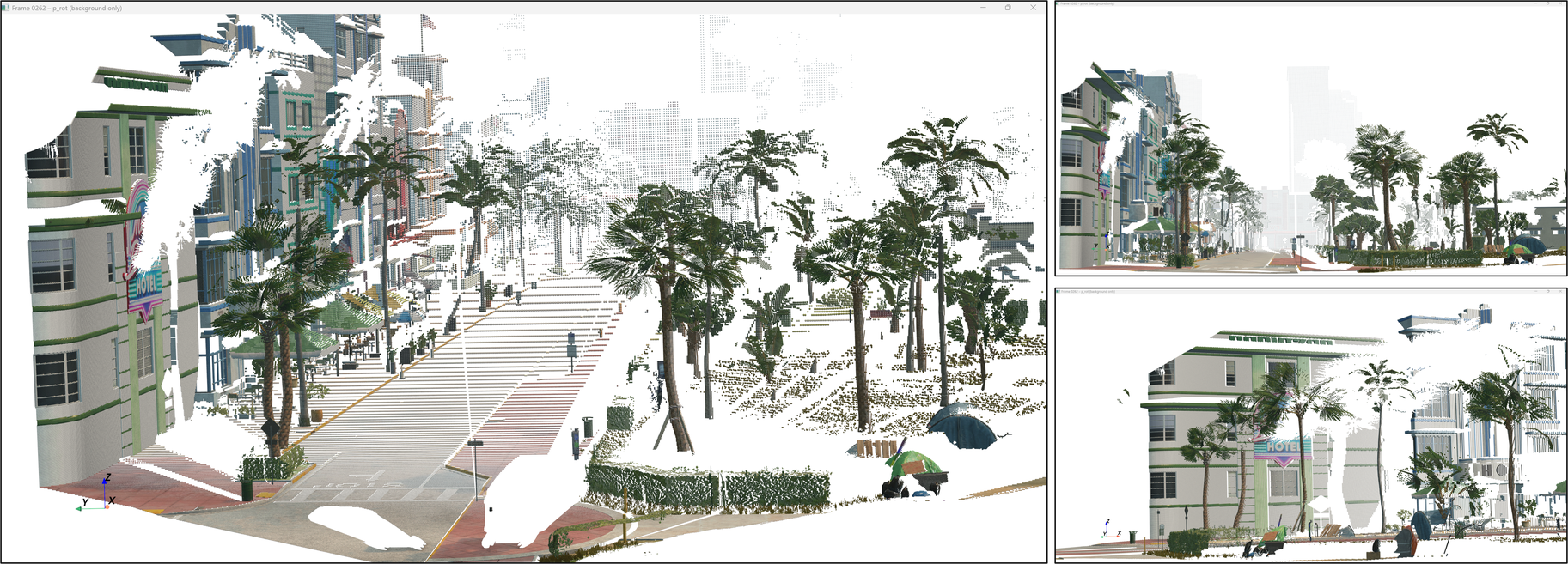}
\caption{Three viewpoints of the RoT point cloud reconstructed for the
stationary scene of Fig.~\ref{fig:stationary_qual}, confirming that the
recovered scaled 3D geometry is consistent with the underlying structure
of the 3D environment.}
\label{fig:stationary_recon}
\end{figure*}

Refer to Figure~\ref{fig:stationary_qual}. With the camera translating through a static environment, the entire
visible scene is a single rigid body, so every image point should recover
the same $\hat{\mathbf{t}}$.
Figure~\ref{fig:stationary_recon} shows reconstruction up to a scale factor using RoT as described in~\cite{raviv2026owl}.

\subsection{Translation-Only Motion}
\begin{figure*}[t]
\centering
\includegraphics[width=0.85\textwidth]{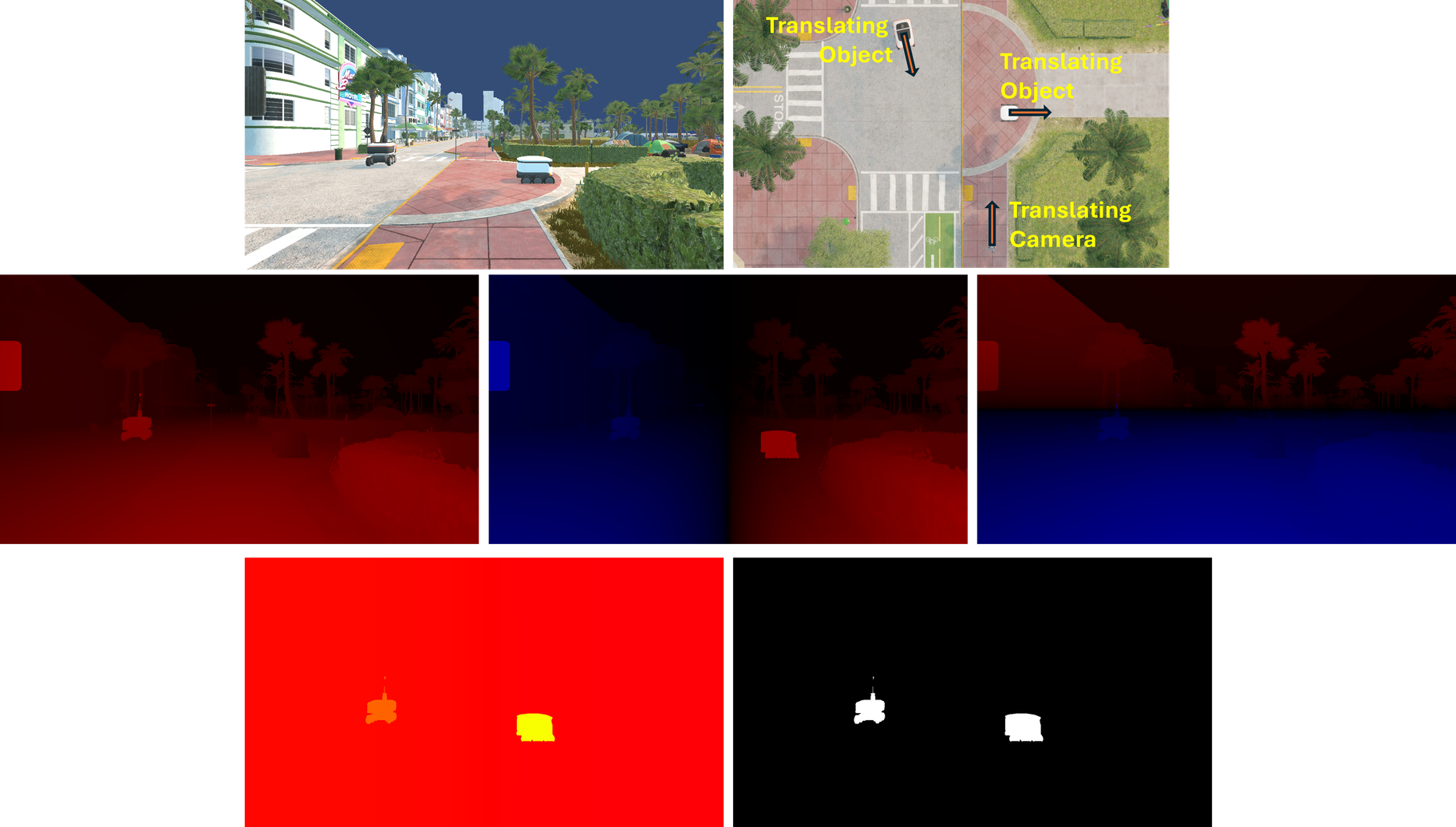}
\caption{Case of translation-only motion. Top row: a representative frame
and a top-down schematic showing two independently translating objects.
Middle row: the corresponding $L$, $\omega_\phi$, and $\omega_\theta$ cues,
in which each object is visible as a localized deviation from the
background. Bottom row: the recovered $\hat{\mathbf{t}}$ field, in which
each object forms its own uniformly colored cluster distinct from the
background, and the corresponding binary object masks. With no self-rotation
on either object, each yields a single well-defined $\hat{\mathbf{t}}$,
cleanly separable by clustering.}
\label{fig:translation_only}
\end{figure*}

Refer to Figure~\ref{fig:translation_only}. Here two rigid bodies undergo pure translation, each with its own distinct
$\hat{\mathbf{t}}$, against the stationary background. These motions are in addition to camera translation.

\subsection{Two Independently Moving Objects: A Degenerate Case}
\begin{figure}[t]
\centering
\includegraphics[width=0.85\columnwidth]{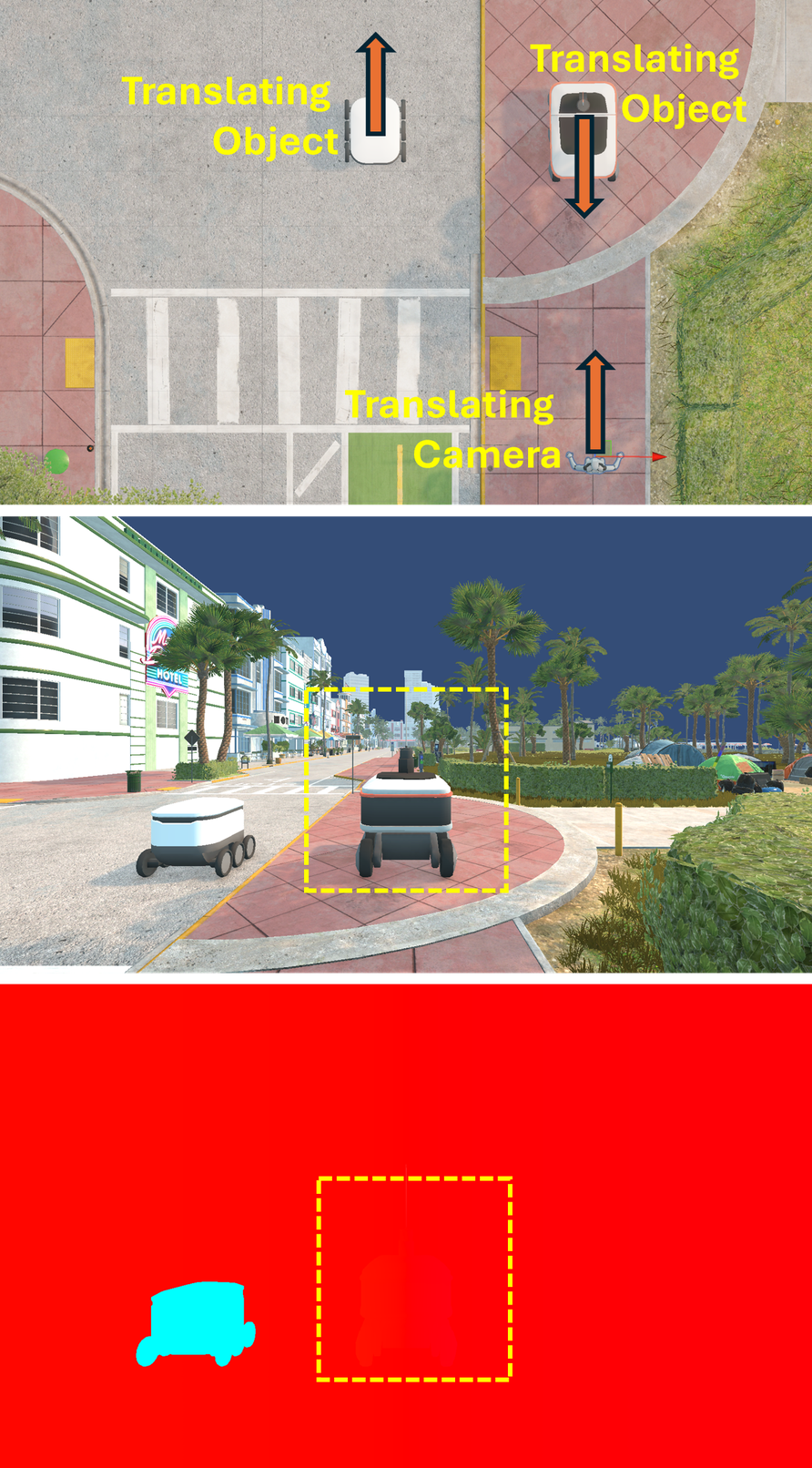}
\caption{Two independently moving objects in a degenerate configuration.
Top: a top-down schematic of the camera and object motions. Middle: a
representative frame, with the robot moving toward the camera boxed.
Bottom: the recovered $\hat{\mathbf{t}}$ field. The boxed robot moves
exactly opposite to the camera, so its $\hat{\mathbf{t}}$ coincides with
that of the background and it does not appear as a separate cluster
(Assumption~3). The second robot forms a distinct cluster.}
\label{fig:two_moving_objects}
\end{figure}

Refer to Figure~\ref{fig:two_moving_objects}. This scenario illustrates the
case excluded by Assumption~3. The camera translates forward while the
boxed robot, in an adjacent lane, moves toward it with exactly the opposite
heading. The camera's velocity relative to that robot therefore points in
the same direction as its velocity relative to the stationary background,
and the two share the same $\boldsymbol{\hat{t}}$. Because
$\boldsymbol{\hat{t}}$ depends only on relative velocity, not on position,
the lateral offset between the lanes does not help. In the
$\boldsymbol{\hat{t}}$ field the boxed robot takes the same color as the
background and cannot be separated by $\boldsymbol{\hat{t}}$ alone. The
second robot, whose relative motion differs, forms a distinct cluster.

\subsection{Self-Rotating Object}
\begin{figure*}[t]
\centering
\includegraphics[width=0.85\textwidth]{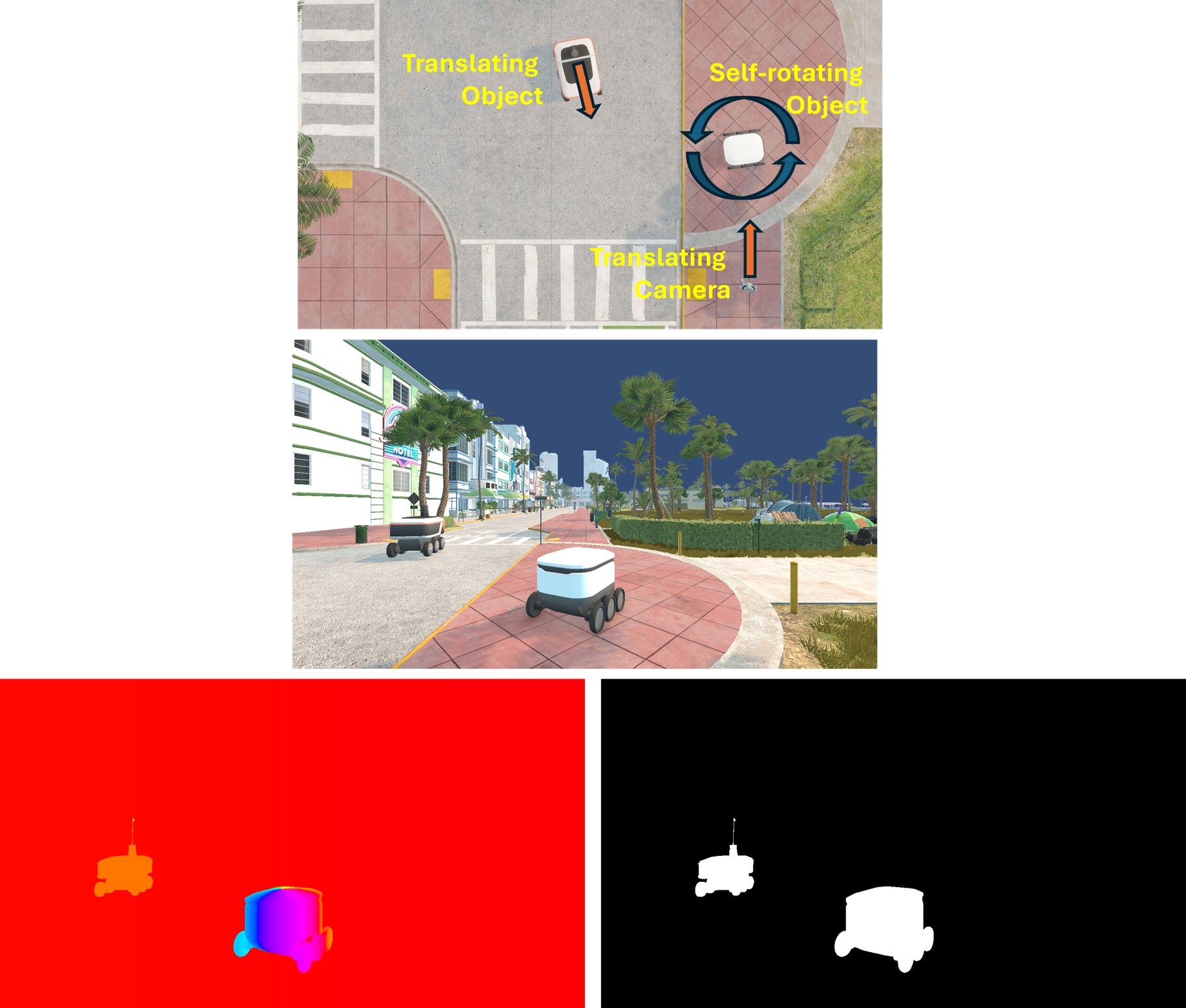}
\caption{Self-rotating object alongside a purely translating object. Top:
a top-down schematic marking the self-rotating object and a representative
frame. Bottom, from left to right: the recovered $\hat{\mathbf{t}}$ field and binary masks.
The purely translating object still yields one uniform
$\hat{\mathbf{t}}$ color, while the self-rotating object's cluster shows
internal color variation, reflecting the local violation of the
pure-translation assumption on that body.}
\label{fig:self_rotating}
\end{figure*}

Refer to Figure~\ref{fig:self_rotating}. Assumption~2 of Section~\ref{sec:method} restricts each rigid body to pure
translation; this scenario tests what happens when that assumption is
violated by one object. It shows one self-rotating object, one purely translating object, and a translating camera.

\subsection{Camera Rotation}
\begin{figure*}[t]
\centering
\includegraphics[width=0.85\textwidth]{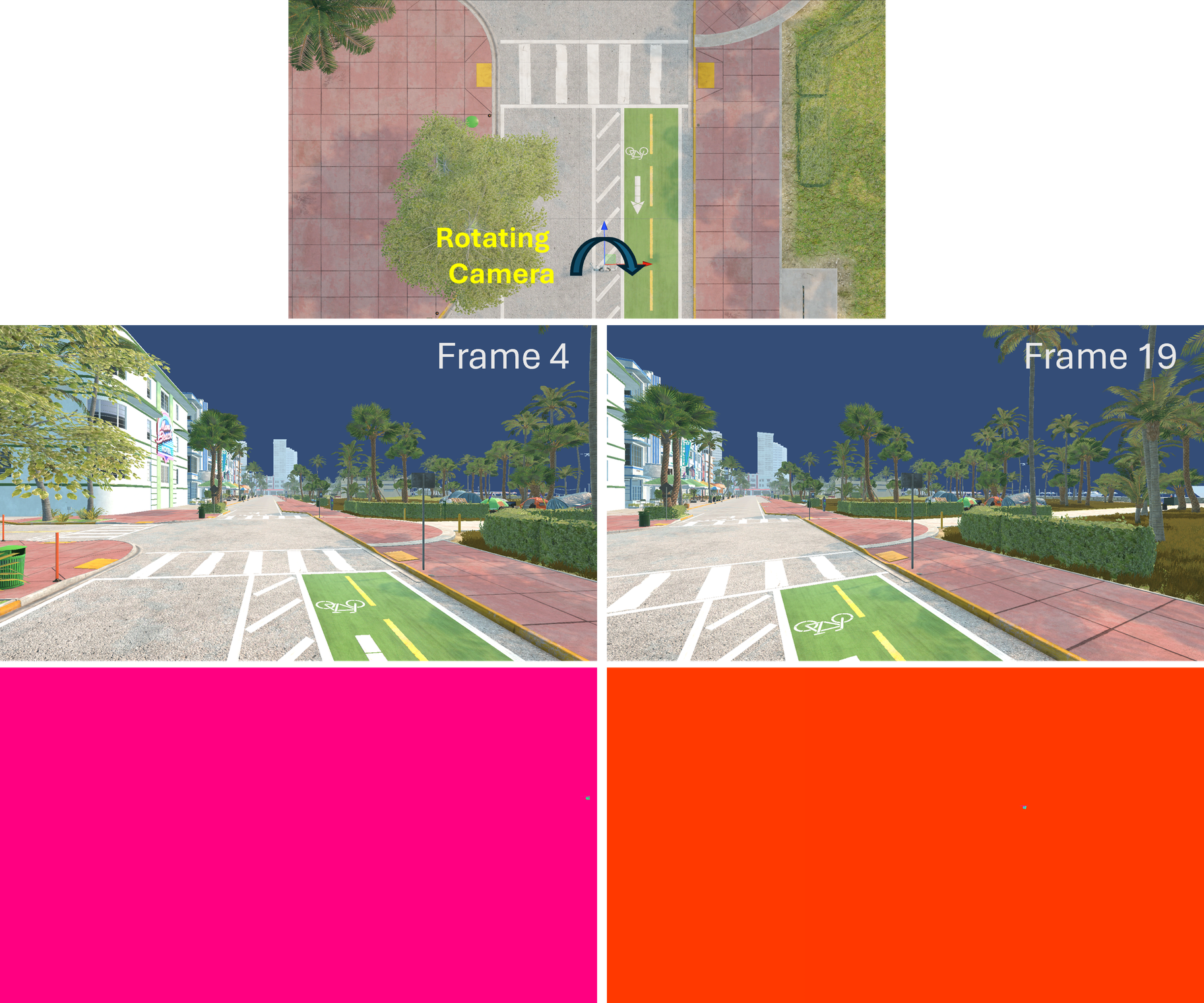}
\caption{Camera rotating, while translating forward, over a stationary
scene. Top: a top-down schematic of the camera's rotation. Bottom: two frames (Frame~4 and Frame~19) sampled
during the rotation, each with its recovered $\hat{\mathbf{t}}$ field shown
below it. In both frames the stationary background still collapses to a
single, uniform $\hat{\mathbf{t}}$ color; the color differs between frames
because $\hat{\mathbf{t}}$ is expressed in the camera's own frame, which has
rotated. Camera rotation does not break the single-cluster structure of the
stationary background.}
\label{fig:camera_rotation}
\end{figure*}

Refer to Figure~\ref{fig:camera_rotation}. This scenario tests the effect of camera rotation on a stationary scene,
and thus the rotation invariance discussed in Section~\ref{sec:seg}.
The camera also translates forward, since~$\boldsymbol{\hat{t}}$ is undefined
when the camera only rotates ($\bt = \mathbf{0}$). The results show that
the single-cluster structure of the stationary background is preserved
under camera rotation.

\subsection{Camera Rotating and Translating: Objects Translating Only}
\begin{figure*}[t]
\centering
\includegraphics[width=0.85\textwidth]{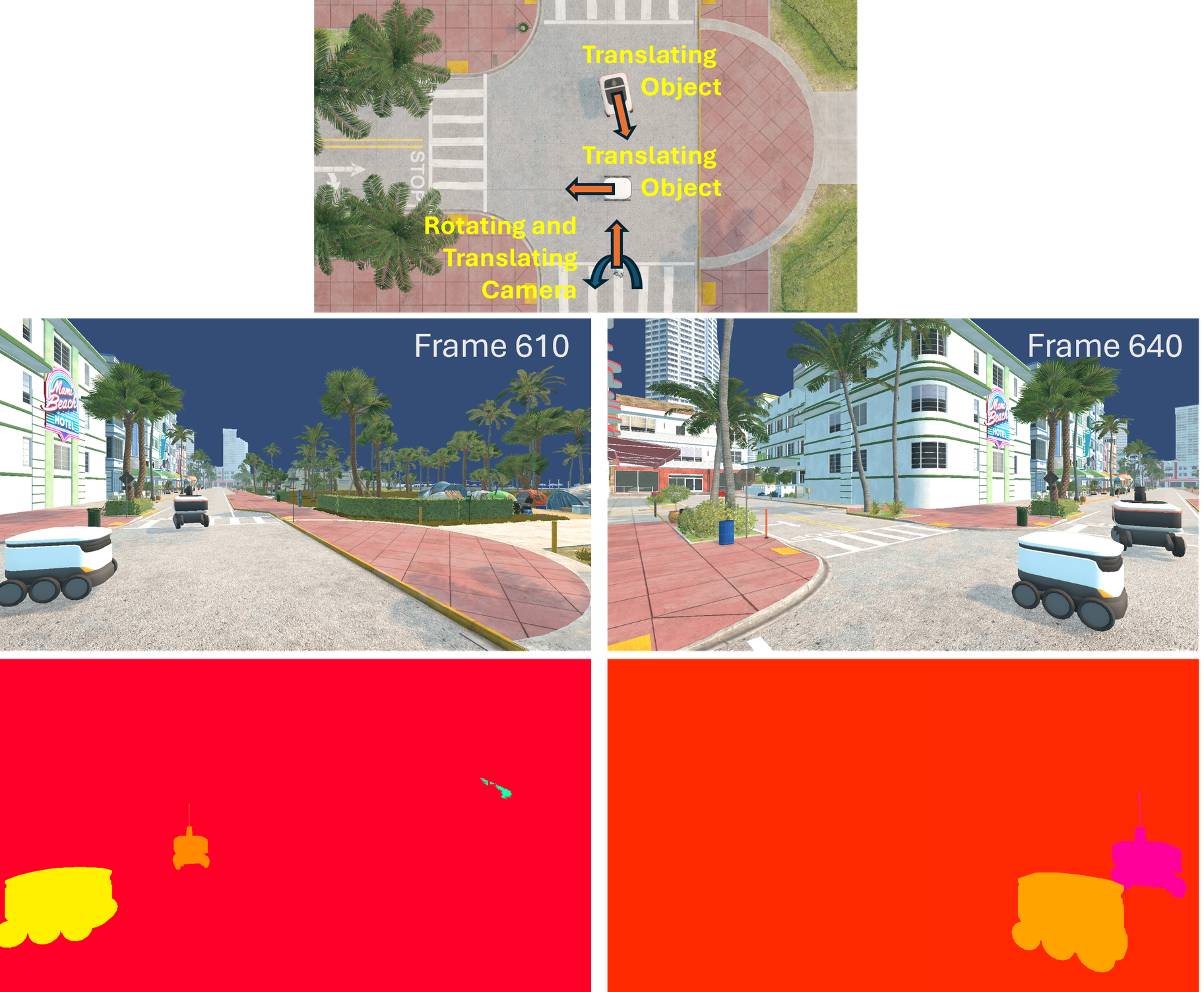}
\caption{Camera rotating and translating, with moving objects translating
only. Top: a top-down schematic of the camera and object motions. Bottom:
two frames (Frame~610 and Frame~640) with their recovered
$\hat{\mathbf{t}}$ fields. The two moving objects remain visible as
distinct, uniformly colored clusters against the background in both
frames, indicating that simultaneous camera rotation and translation does
not disrupt the recovery of per-object $\hat{\mathbf{t}}$.}
\label{fig:cam_rot_trans_translate_only}
\end{figure*}

Refer to Figure~\ref{fig:cam_rot_trans_translate_only}. This scenario shows the effect of camera translating and rotating relative to translating objects.

\subsection{Camera Rotating and Translating: Objects Translating and Self-Rotating}
\begin{figure*}[t]
\centering
\includegraphics[width=0.85\textwidth]{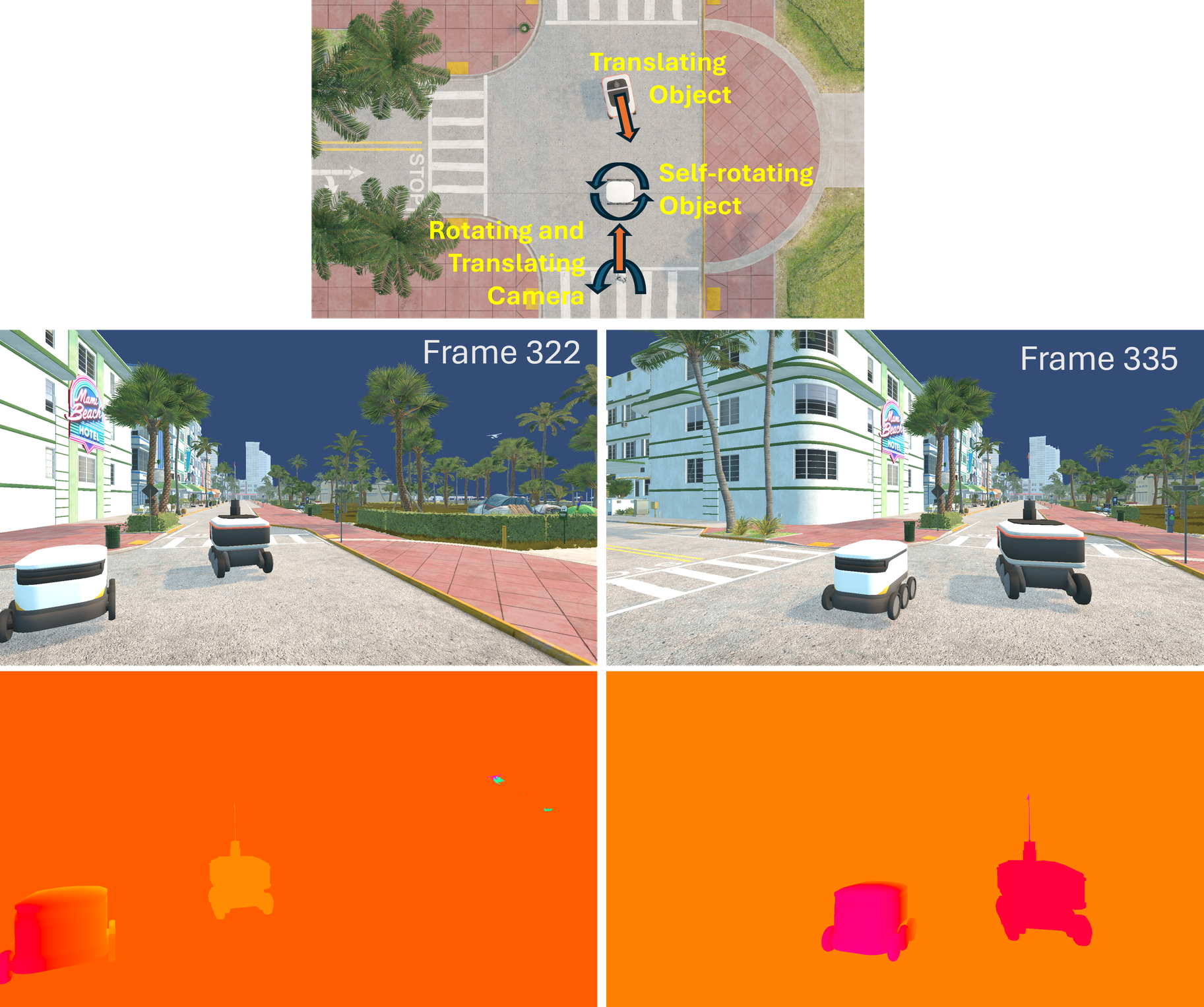}
\caption{Camera rotating and translating, with two moving objects, the one on the right is translating only and the one on the left rotates only. Top: a top-down schematic of the combined motions at an earlier frame than Frame 322.
Bottom: two frames (Frame~322 and Frame~335) with their recovered
$\hat{\mathbf{t}}$ fields. As in Fig.~\ref{fig:self_rotating}, the
self-rotation of the objects introduces internal color variation within
their clusters, now superimposed on the effects of camera rotation and
translation.}
\label{fig:cam_rot_trans_translate_rotate}
\end{figure*}

Refer to Figure~\ref{fig:cam_rot_trans_translate_rotate}. In this case the camera translates and rotates, while one object self-rotates and another translates.

\subsection{A More General Case}
\begin{figure}[t]
\centering
\includegraphics[width=0.9\columnwidth]{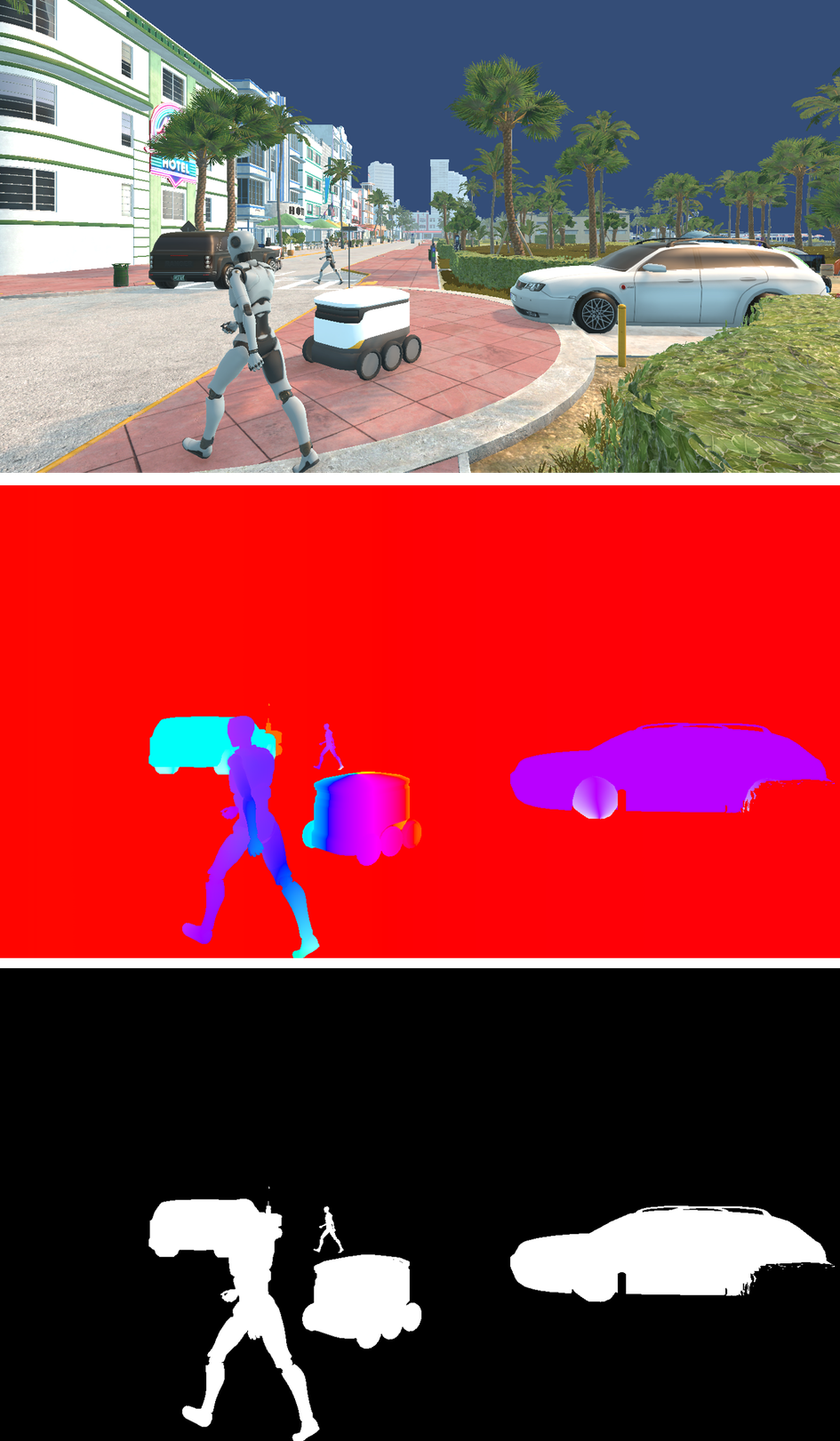}
\caption{A more general case: multiple independently moving objects (two
cars, a walking humanoid robot, a delivery robot, and a pedestrian)
together with camera motion. Top: a representative frame. Middle: the
recovered $\hat{\mathbf{t}}$ field, in which each moving object stands out
from the uniform background. Bottom: the corresponding binary masks of
moving versus static regions. The rigid vehicles produce nearly uniform
$\hat{\mathbf{t}}$, while the articulated humanoid and pedestrian, which
violate the rigid-body assumption, show internal $\hat{\mathbf{t}}$
variation but remain clearly distinct from the background.}
\label{fig:general_case}
\end{figure}

Refer to Figure~\ref{fig:general_case}. The general case combines multiple independently moving objects with
camera translating and rotating.

\section{Conclusions}

This paper has extended the \textit{OWL} perception framework~\cite{raviv2026owl}
to the problem of segmenting multiple independently moving rigid objects from
visual motion cues observed by a single camera. The central analytical result is
that the unit translation direction $\hat{\mathbf{t}}$, recoverable per image
point from the perceived looming $L$ and perceived rotation $\boldsymbol{\omega}$,
is a signature of rigid body membership: all points on the same non-rotating
rigid body share a common $\hat{\mathbf{t}}$, while points on distinct bodies
yield different values, except in the degenerate configuration excluded by
Assumption~3.

The resulting framework is geometrically principled, depth-free, training-free,
and fully parallel. It exploits the quaternion structure of \textit{OWL} to
recover $\hat{\mathbf{t}}$ from a single point when the angular velocity vector
$\boldsymbol{\omega}$ is available, as it is from de-rotated optical flow; when
only the magnitude $\omega$ is known, three points are needed, which gives an
intuitive picture of $\hat{\mathbf{t}}$ as the common direction of three cones.
Camera rotation is compensated by construction through the de-rotation identity
for $\boldsymbol{\omega}$. Once segmented, each cluster can be passed
independently to the \textit{OWL}/RoT reconstruction of~\cite{raviv2026owl},
which would yield a scaled 3D point cloud per body, each with its own scale
factor $1/|\mathbf{t}_k|$. Implementing this per-segment reconstruction is
ongoing work.

The present work focuses on the instantaneous derivation of $\hat{\mathbf{t}}$
from a single time instant. An important direction for future work is to observe
$\hat{\mathbf{t}}$ over time: tracking how $\hat{\mathbf{t}}$ evolves across
frames will enable more robust segmentation, noise rejection, predictions and the
detection of objects that change their direction of motion. This temporal
extension, together with integration with foundation-model pipelines such as
SAM~2~\cite{ravi2024sam2}, can be the focus of subsequent work.

Together with~\cite{raviv2026owl}, this paper forms the second element of a
unified, perception-grounded pipeline: the \textit{OWL} paper~\cite{raviv2026owl}
establishes scaled 3D reconstruction and heading recovery, and the present paper
establishes multi-body segmentation as a companion step. Applying \textit{OWL}/RoT
independently to each segment, the subject of ongoing work, completes the
pipeline. Both steps require no training data and no depth sensors; they need
only the camera intrinsics and the camera's rotation rate.

\appendices

\section{Vector Derivation of \texorpdfstring{$\hat{\mathbf{t}}$}{t-hat} Recovery}
\label{app:vector_derivation}

This appendix derives the unit translation direction~$\hat{\mathbf{t}}$
using only standard 3-D vector algebra, without invoking quaternion
arithmetic.  The result is identical to~\eqref{eq:tcap_explicit}; the
purpose here is to make the geometric content transparent for readers
more familiar with vector notation.

\subsection*{Setup}

Let $\hat{e}_r$ be the unit line-of-sight vector from the camera to an
image point, and let $\mathbf{t}$ be the translational velocity of the camera
relative to the observed rigid body, as in~\cite{raviv2026owl}.  The two OWL cues measured at that pixel
are:
\begin{align}
  L      &= \frac{\mathbf{t}\cdot\hat{e}_r}{|\mathbf{r}|},
  \label{eq:app_L} \\
  \boldsymbol{\omega} &= \frac{\hat{e}_r\times\mathbf{t}}{|\mathbf{r}|}.
  \label{eq:app_w}
\end{align}
$L$~is a scalar (the looming); $\boldsymbol{\omega}$~is a vector
perpendicular to~$\hat{e}_r$ (the perceived rotation).
Both are directly observable from the image sequence.
The goal is to recover $\hat{\mathbf{t}} = \mathbf{t}/|\mathbf{t}|$
from $L$, $\boldsymbol{\omega}$, and $\hat{e}_r$ alone.

\subsection*{Step 1: Decompose \texorpdfstring{$\mathbf{t}$}{t} into radial and tangential parts}

Any vector $\mathbf{t}$ can be split into a component \emph{along}
$\hat{e}_r$ and a component \emph{perpendicular} to it:
\begin{equation}
  \mathbf{t}
    = \underbrace{(\mathbf{t}\cdot\hat{e}_r)\,\hat{e}_r}_{\text{radial}}
    + \underbrace{\mathbf{t} - (\mathbf{t}\cdot\hat{e}_r)\,\hat{e}_r}_{\text{tangential}}.
  \label{eq:app_decomp}
\end{equation}

From~\eqref{eq:app_L}, the radial part (divided by $|\mathbf{r}|$)
is immediately:
\begin{equation}
  \frac{(\mathbf{t}\cdot\hat{e}_r)\,\hat{e}_r}{|\mathbf{r}|}
    = L\,\hat{e}_r.
  \label{eq:app_radial}
\end{equation}

\subsection*{Step 2: Recover the tangential part via a second cross product}

The tangential part of $\mathbf{t}$ is not directly visible in
$\boldsymbol{\omega}$, but it can be extracted by crossing
$\hat{e}_r$ with $\boldsymbol{\omega}$.
Substituting~\eqref{eq:app_w}:
\begin{equation}
  \hat{e}_r\times\boldsymbol{\omega}
    = \hat{e}_r\times\frac{(\hat{e}_r\times\mathbf{t})}{|\mathbf{r}|}.
  \label{eq:app_cross}
\end{equation}

Apply the \textbf{vector triple product identity}
$\mathbf{a}\times(\mathbf{a}\times\mathbf{b})
 = \mathbf{a}(\mathbf{a}\cdot\mathbf{b})-\mathbf{b}(\mathbf{a}\cdot\mathbf{a})$
with $\mathbf{a}=\hat{e}_r$ (so $\hat{e}_r\cdot\hat{e}_r=1$)
and $\mathbf{b}=\mathbf{t}$:
\begin{equation}
  \hat{e}_r\times(\hat{e}_r\times\mathbf{t})
    = (\mathbf{t}\cdot\hat{e}_r)\,\hat{e}_r - \mathbf{t}.
  \label{eq:app_ttp}
\end{equation}

Inserting~\eqref{eq:app_ttp} into~\eqref{eq:app_cross} and
rearranging:
\begin{equation}
  \frac{\mathbf{t}}{|\mathbf{r}|}
    = \frac{(\mathbf{t}\cdot\hat{e}_r)\,\hat{e}_r}{|\mathbf{r}|}
      - \hat{e}_r\times\boldsymbol{\omega}
    = L\,\hat{e}_r - \hat{e}_r\times\boldsymbol{\omega}.
  \label{eq:app_t_over_r}
\end{equation}

\subsection*{Step 3: Normalize}

Equation~\eqref{eq:app_t_over_r} expresses $\mathbf{t}/|\mathbf{r}|$
entirely in terms of observables.
Its magnitude is $|\mathbf{t}|/|\mathbf{r}| = \sqrt{L^2+\omega^2}$
(the OWL speed-to-range ratio), so dividing gives the unit vector:
\begin{equation}
  \boxed{
    \hat{\mathbf{t}}
      = \frac{L\,\hat{e}_r \;-\; \hat{e}_r\times\boldsymbol{\omega}}
             {\sqrt{L^2+\omega^2}}
  }.
  \label{eq:app_tcap}
\end{equation}

This is identical to~\eqref{eq:tcap_explicit}.
No scene depth, or knowledge of $|\mathbf{t}|$ or $|\mathbf{r}|$
individually, is required.

\subsection*{Physical interpretation}

\begin{figure}[t]
\centering
\begin{tikzpicture}[>=Stealth, thick, scale=1.25]
  \coordinate (O) at (0,0);

  \draw[->, gray!60, line width=1pt] (O) -- (3.4,0)
    node[right, gray] {$\hat{e}_r$};

  \draw[->, blue!80!black, line width=1.8pt] (O) -- (2.6,1.8)
    node[above right] {$\mathbf{t}/|\mathbf{r}|$};

  \draw[->, red!70!black, line width=1.5pt] (O) -- (2.6,0)
    node[below] {$L\,\hat{e}_r$};

  \draw[->, orange!80!black, line width=1.5pt] (2.6,0) -- (2.6,1.8)
    node[right, xshift=2pt] {$-\hat{e}_r\!\times\!\boldsymbol{\omega}$};

  \draw (2.6,0.22) -- (2.38,0.22) -- (2.38,0);

  \draw[gray] (0.9,0) arc[start angle=0, end angle=34.7, radius=0.9]
    node[midway, right, font=\small, gray] {$\alpha$};

\end{tikzpicture}
\caption{%
  Geometric decomposition of $\mathbf{t}/|\mathbf{r}|$ in the plane
  containing $\mathbf{t}$ and $\hat{e}_r$.
  The radial component $L\,\hat{e}_r$ (red) is read directly from the
  looming.  The tangential component $-\hat{e}_r\times\boldsymbol{\omega}$
  (orange) is recovered by a second cross product.
  Their vector sum (blue) gives $\mathbf{t}/|\mathbf{r}|$;
  normalizing by $\sqrt{L^2+\omega^2}$ yields $\hat{\mathbf{t}}$.}
\label{fig:app_decomp}
\end{figure}
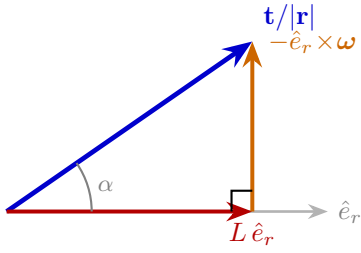

Each term in the numerator of~\eqref{eq:app_tcap} has a direct
geometric meaning (see Fig.~\ref{fig:app_decomp}):

\begin{itemize}
  \item $L\,\hat{e}_r$: the component of $\mathbf{t}$ \emph{along}
        the line of sight, read directly from the looming scalar~$L$.
  \item $-\hat{e}_r\times\boldsymbol{\omega}$: the component of
        $\mathbf{t}$ \emph{across} the line of sight, recovered by
        un-doing the cross product that created $\boldsymbol{\omega}$
        in the first place.
\end{itemize}

Adding these two orthogonal pieces reconstructs $\mathbf{t}$ in
direction, and dividing by $\sqrt{L^2+\omega^2}$ normalizes to unit
length.  The quaternion multiplication in the main text performs
exactly these two operations in a single algebraic step via the
quaternion product rule; the derivation above makes the underlying
vector geometry explicit.

\bibliographystyle{IEEEtran}
\bibliography{raviv-yepes}

\end{document}